\documentclass[conference]{IEEEtran}
\IEEEoverridecommandlockouts

\usepackage{cite}
\usepackage{amsmath,amssymb,amsfonts}
\usepackage{algorithmic}
\usepackage{graphicx}
\usepackage{textcomp}
\usepackage{xcolor}
\def\BibTeX{{\rm B\kern-.05em{\sc i\kern-.025em b}\kern-.08em
    T\kern-.1667em\lower.7ex\hbox{E}\kern-.125emX}}
\begin{document}

\title{Efficient Edge-Compatible CNN for Speckle-Based Material Recognition in Laser Cutting Systems

}

\author{\IEEEauthorblockN{Mohamed Abdallah Salem}
\IEEEauthorblockA{\textit{College of Engineering} \\
\textit{North Dakota State University}\\
Fargo, ND, USA, 58102 \\
mohamed-salem@ieee.org}
\and
\IEEEauthorblockN{Nourhan Zein Diab}
\IEEEauthorblockA{\textit{College of Computer Science and Engineering} \\
\textit{New Mansoura University}\\
New Masoura, Egypt \\
nourdiab736@gmail.com}
% \and
}

\maketitle

\begin{abstract}
Accurate material recognition is critical for safe and effective laser cutting, as misidentification can lead to poor cut quality, machine damage, or the release of hazardous fumes. Laser speckle sensing has recently emerged as a low-cost and non-destructive modality for material classification; however, prior work has either relied on computationally expensive backbone networks or addressed only limited subsets of materials. In this study, A lightweight convolutional neural network (CNN) tailored for speckle patterns is proposed, designed to minimize parameters while maintaining high discriminative power. Using the complete SensiCut dataset of 59 material classes spanning woods, acrylics, composites, textiles, metals, and paper-based products, the proposed model achieves 95.05\% test accuracy, with macro and weighted F1-scores of 0.951. The network contains only 341k trainable parameters ($\sim$1.3 MB)---over 70$\times$ fewer than ResNet-50---and achieves an inference speed of 295 images per second, enabling deployment on Raspberry Pi and Jetson-class devices. Furthermore, when materials are regrouped into nine and five practical families, recall exceeds 98\% and approaches 100\%, directly supporting power and speed preset selection in laser cutters. These results demonstrate that compact, domain-specific CNNs can outperform large backbones for speckle-based material classification, advancing the feasibility of material-aware, edge-deployable laser cutting systems.
\end{abstract}

\begin{IEEEkeywords}
Laser speckle, Material recognition, Lightweight CNN, Deep learning, Edge AI, SensiCut dataset
\end{IEEEkeywords}

\section{Introduction}

Laser cutting has become an indispensable process in modern manufacturing, makerspaces, and prototyping environments due to its precision, flexibility, and ability to handle a wide range of materials. A focused, high-energy laser beam enables rapid cutting, engraving, or marking across materials such as wood, acrylic, polymers, textiles, and metals. However, each material interacts differently with laser energy depending on its composition, surface reflectivity, and thermal properties. As a result, correct material identification prior to cutting is essential to ensure clean edges, minimize burn marks, and prevent safety hazards \cite{rodrigues2017pvc,barnes2015polycarbonate}. For example, polyvinyl chloride (PVC) releases corrosive hydrochloric acid and chlorine gas when lasered, while polycarbonate (Lexan) tends to melt or discolor under standard CO$_2$ laser settings \cite{barnes2015polycarbonate}. Improper identification can therefore lead to suboptimal quality, machine damage, or exposure to toxic fumes.

Traditional laser cutters rely on user input or manual preset selection to specify the material type and associated power, speed, and frequency parameters. Although effective in controlled environments, these manual methods are prone to human error-especially in shared or automated systems-resulting in inconsistent results and potential safety risks. The growing use of autonomous fabrication tools in educational, industrial, and research settings calls for intelligent systems that can automatically recognize materials and adjust cutting parameters accordingly.

Recent advances in optical sensing have made this goal achievable. One of the most promising modalities for non-contact material recognition is laser speckle sensing, which leverages the unique interference patterns produced when coherent light interacts with a material's microstructure. These speckle patterns act as intrinsic optical fingerprints, encoding information about the material's surface and internal properties. When combined with modern deep learning algorithms, speckle imaging enables precise, rapid, and non-destructive classification of materials without any physical contact or surface preparation \cite{dogan2021sensicut, castilho2023machine, shimadera2022speckle, yan2023lasershoes}. 

Laser speckle sensing has emerged as a promising solution for material classification. Speckle patterns arise from the interference of coherent light scattered by a rough surface and serve as unique fingerprints reflecting the microstructure of the material \cite{duque2024experimental}. This sensing modality is non-destructive, requires minimal hardware (laser diode and camera), and is robust under varying orientations. Beyond fabrication, speckle imaging has been widely utilized in biomedical optics for blood flow monitoring and tissue characterization \cite{boas2010speckle} and in agriculture for fruit quality assessment and defect detection \cite{singh2020fruit}, underscoring its versatility as a low-cost and powerful technique.

Despite its potential, deploying deep learning models for speckle-based classification in laser cutters presents several challenges. Most high-performing convolutional neural networks (CNNs) are computationally expensive, requiring large memory and processing power that exceed the capacity of embedded systems typically used in desktop laser cutters. Moreover, compact architectures must still maintain discriminative capability to distinguish visually similar materials such as MDF, plywood, and tinted acrylics. Consequently, there is a growing need for lightweight, edge-deployable neural networks that can deliver both high accuracy and real-time performance on low-power hardware.

To address this need, this study develops an efficient, lightweight CNN optimized for speckle-based material recognition in laser cutting systems. The model is designed to minimize computational complexity while retaining high discriminative power across a comprehensive material dataset encompassing 59 distinct classes. The resulting network enables reliable material identification and direct mapping to laser cutter power and speed presets, paving the way for safer, more autonomous, and material-aware fabrication systems.

\section{Related Work}

The first major demonstration of speckle-based material recognition for laser cutting was introduced by Dogan \textit{et al.} \cite{dogan2021sensicut} in the \emph{SensiCut} system. A compact green laser and camera module were mounted to the laser cutter head, and a dataset was collected covering 30 material types (59 subclasses). A ResNet-50 model trained via transfer learning achieved over 98\% accuracy on 30 classes. However, performance dropped to approximately 94\% when scaling to the complete 59-class dataset, and the 25M-parameter backbone limited feasibility for deployment on resource-constrained systems. The study highlighted the feasibility of speckle-based recognition but exposed challenges in scalability and efficiency.

Salem \textit{et al.} \cite{salem2023material} expanded this direction by constructing a dataset of 6,000 speckle images across 20 materials, including ceramics, marble, thermoplastic polyurethane (TPU), and layered samples with protective coatings. Using VGG16, they achieved 98.1\% accuracy and demonstrated robustness across orientation and illumination conditions. While this work addressed practical issues such as protective films and engraving materials, its reliance on a large backbone (VGG16, $\sim$138M parameters) restricted applicability in real-time fabrication settings. This underscores the need for compact models that can operate under practical deployment constraints.

In subsequent work, Salem \textit{et al.} \cite{salem2023detection} investigated hazardous material detection, focusing on PVC, Lexan, ABS, and carbon fiber. The authors proposed a custom CNN trained on the SensiCut dataset with single-channel (laser-matched) input, achieving 97.8\% accuracy and a macro F1-score of 0.978. This approach reduced inference time by 86.5\% compared to ResNet-50 while retaining accuracy, highlighting the potential of lightweight architectures for safety-critical applications. However, the study remained limited to hazardous subsets rather than the complete material spectrum, leaving open the challenge of comprehensive classification.

In parallel with research on speckle-based material recognition, the computer-vision community has witnessed significant advances in compact convolutional neural network (CNN) architectures tailored for edge deployment. Surveys of edge-friendly neural networks highlight that fewer than half of the studied models have been validated on real hardware platforms such as ARM-based processors or embedded devices \cite{wang2025review}. For example, the RepViT family revisits lightweight CNN design from the perspective of Vision Transformers: Wang \textit{et al.} demonstrated that by gradually infusing ViT-inspired architectural features into MobileNetV3\cite{howard2017mobilenets}, a pure CNN backbone could achieve over 80\% top-1 accuracy on ImageNet while attaining approximately 1 ms latency on an iPhone 12 \cite{wang2023repvit}. Another recent architecture, EfficientNet-eLite, targets hardware-friendliness explicitly by scaling down both input resolution and network width/depth, achieving improved parameter efficiency compared to existing compact models \cite{wang2022efficientnetelite}. 

Meanwhile, RapidNet proposes a purely CNN-based mobile backbone using multi-level dilated convolutions and reports a top-1 accuracy of 76.3\% on ImageNet at 0.9 ms latency on an iPhone 13 mini NPU \cite{munir2024rapidnet}. These developments illustrate that domain-agnostic lightweight CNNs are rapidly closing the performance gap while meeting stringent latency and memory constraints of edge hardware. However, few of these architectures have been applied to speckle-pattern recognition or integrated into real-time machine-control systems such as laser cutters. This gap underlines the need for a domain-tailored, ultra-compact network that not only handles fine-grained classification across many classes but also supports actual edge deployment in fabrication environments.

The significance of this research lies in bridging the gap between high-accuracy laboratory demonstrations and practical, real-world deployment of speckle-based material recognition in laser cutters. Unlike prior works that either:  
\begin{itemize}
    \item Focused on limited subsets of the dataset \cite{dogan2021sensicut},  
    \item Relied on heavy backbone architectures \cite{salem2023material}, or  
    \item Restricted scope to hazardous materials only \cite{salem2023detection},  
\end{itemize}

The present study demonstrates that a domain-tailored, lightweight CNN can scale to the complete 59-class SensiCut dataset while maintaining high accuracy. Specifically, the proposed model achieves 95.05\% accuracy and a macro F1-score of 0.951 using only 341k parameters ($\sim$1.3 MB), making it over 70$\times$ smaller than ResNet-50. The inference speed of $\sim$295 images enables real-time classification on Raspberry Pi or Jetson-class edge devices.

Furthermore, grouping materials into five and nine families was found to result in near-perfect recognition($\geq$98\% recall), directly mapping to laser cutter presets for power, speed, and frequency. This contribution is particularly impactful as it integrates recognition with machine control, improving both safety (e.g., PVC detection) and efficiency (e.g., optimized presets for wood, acrylic, paper). Thus, this study positions lightweight CNNs as practical, deployable successors to heavyweight architectures, advancing material-aware laser cutting and opening new directions for speckle-based sensing in fabrication and beyond.

The remainder of this paper is organized as follows. Section~\ref{sec:dataset} details the dataset composition, preprocessing steps, and augmentation procedures. Section~\ref{sec:architecture} describes the design of the proposed lightweight CNN and its architectural rationale. Section~\ref{sec:evaluation} presents the experimental setup, performance analysis, and comparisons with related architectures. Section~\ref{sec:discussion} discusses the implications of the results and outlines key insights. Finally, Section~\ref{sec:conclusion} concludes the paper and proposes directions for future work.

\section{Dataset and Preprocessing}
\label{sec:dataset}
This study employs the SensiCut dataset, which is one of the most comprehensive publicly available datasets for speckle-based material recognition. The dataset contains speckle images captured from 59 different material classes, each with multiple samples per class to account for variations in orientation, illumination, and surface texture. The materials span a wide range of categories relevant to laser cutting, including natural and engineered woods (e.g., hardwoods, MDF variants, plywood), a diverse set of acrylics (cast, matte, extruded, tinted, and colored), polymers such as Delrin, ABS, and PVC, composites like Lexan and carbon fiber, textiles and soft materials such as felt, suede, leather, and wool, and common paper-based products such as cardstock, corrugated cardboard, and coasterboard. The inclusion of metals such as stainless steel and aluminum further increases the diversity of the dataset, making it suitable for evaluating real-world cutting scenarios.

To prepare the dataset for training, each speckle image was preprocessed by extracting only the green channel. Prior studies have shown that the green channel preserves speckle contrast more effectively than red or blue, which makes it a suitable choice for single-channel training while reducing computational overhead. The images were converted to grayscale using this channel, rescaled to the [0,1] range, and resized to $512 \times 512$ pixels to standardize the input dimension. To improve generalization and increase robustness to orientation and surface variability, data augmentation was applied in the form of horizontal and vertical flips. The dataset was divided into training, validation, and test sets, with the test set comprising 364 images evenly distributed across the 59 material classes. This preprocessing pipeline ensured that the input data maintained high speckle fidelity while remaining computationally efficient.

\begin{figure}[htbp]
    \centering
    \includegraphics[width=\columnwidth]{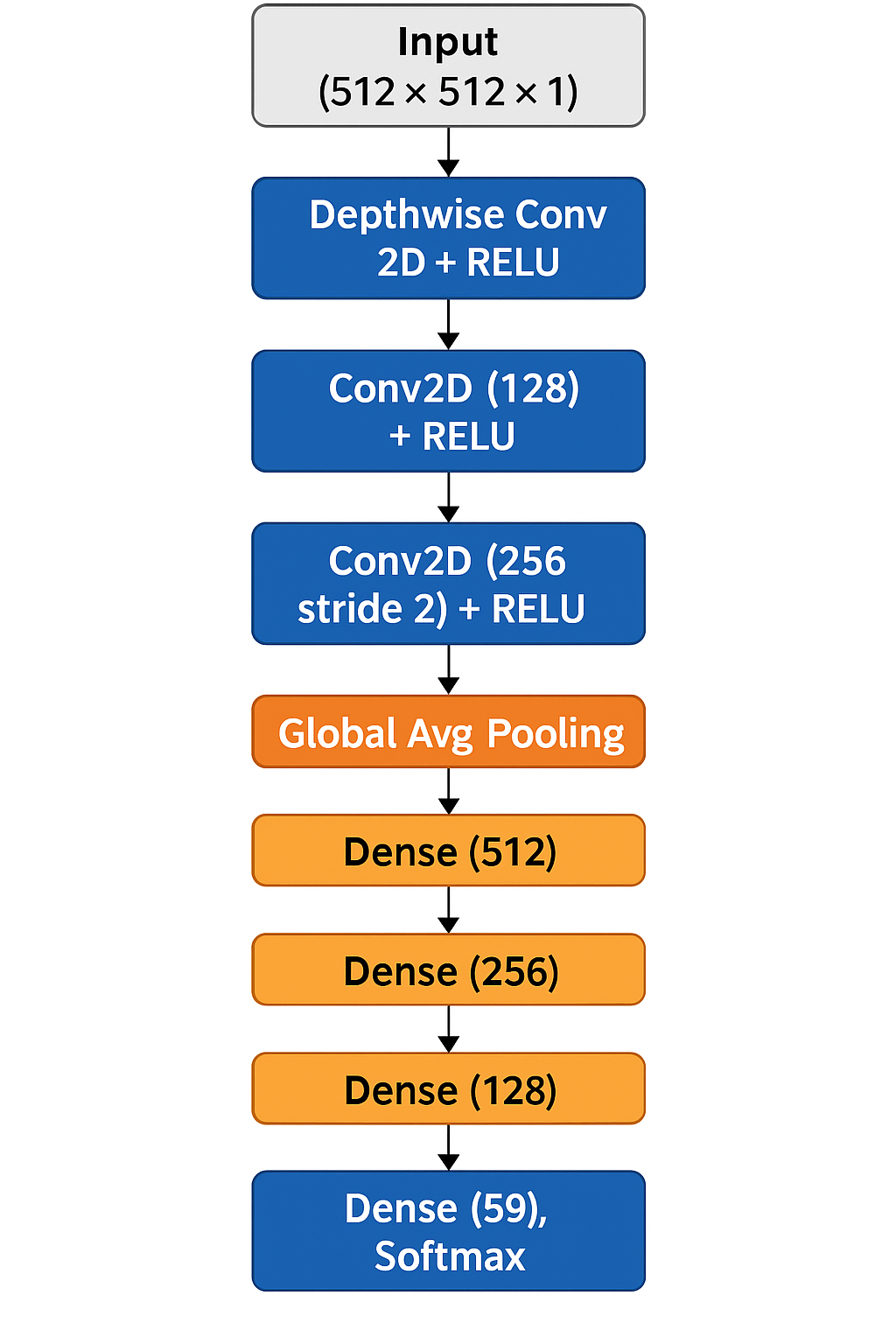}
% \centerline{\width=\includegraphics{Figs/Model_Arc.png}}
\caption{Proposed lightweight CNN architecture for speckle-based material classification. The design balances accuracy and efficiency using depthwise and pointwise convolutions, global average pooling, and fully connected layers.}
\label{fig:Model}
\end{figure}

\section{Proposed CNN Architecture}
\label{sec:architecture}
The proposed CNN architecture was specifically designed to balance accuracy with efficiency, aiming to minimize the number of parameters while retaining discriminative power for fine-grained speckle classification. As illustrated in Fig.~\ref{fig:Model}, the model accepts a $512 \times 512$ single-channel input. The first convolutional layer applies 32 filters of size $3 \times 3$ with a stride of 2, effectively reducing the spatial resolution while capturing low-level texture features. A Rectified Linear Unit (ReLU) activation follows to introduce non-linearity. 

To further reduce computational cost, a Depthwise Convolution is applied next. This layer performs channel-wise filtering, significantly lowering the parameter count compared to standard convolutions. The resulting feature maps are then passed to a $1 \times 1$ convolution with 128 filters, which serves as a pointwise operation to recombine the depthwise outputs and enhance feature diversity. Another $1 \times 1$ convolution with 256 filters and stride 2 follows, introducing additional downsampling and capturing higher-level abstract features while maintaining efficiency. Each convolutional block is paired with a ReLU activation to ensure non-linear feature transformation.

After the convolutional stages, a Global Average Pooling (GAP) layer is employed to replace traditional flattening. This design reduces the number of trainable parameters while aggregating spatial features into compact descriptors, which improves both generalization and robustness. The pooled features are fed into three fully connected layers of sizes 512, 256, and 128 neurons, respectively, each activated by ReLU functions. The final classification layer consists of 59 neurons with a Softmax activation, corresponding to the 59 material classes. 

Despite its simplicity, the network achieves strong discriminative ability with only 341,307 trainable parameters, equivalent to approximately 1.3 MB of memory. This compact footprint makes the model suitable for edge deployment without sacrificing classification performance.

\subsection{Comparison with Related Architectures}
The design of the proposed CNN, illustrated in Fig.~\ref{fig:Model}, differs substantially from previously employed backbones such as ResNet-50 \cite{dogan2021sensicut} and VGG16 \cite{salem2023material}. Unlike these deep, multi-stage architectures that rely on numerous convolutional blocks and residual connections, the proposed model employs a shallow feature extraction pipeline composed of an initial $3 \times 3$ convolution, a depthwise convolution, and two subsequent $1 \times 1$ pointwise convolutions. This configuration significantly reduces the number of parameters and floating-point operations while retaining the model's ability to capture the fine-grained texture variations inherent to laser speckle patterns. The inclusion of depthwise and pointwise convolutions follows the principles of MobileNet \cite{howard2017mobilenets} and ShuffleNet \cite{zhang2018shufflenet}, but the overall structure is adapted to single-channel speckle inputs and does not rely on group convolutions or channel shuffling, further simplifying deployment.

The adoption of global average pooling (GAP) instead of flattening or fully connected layers minimizes overfitting and contributes to a dramatic reduction in parameters compared to VGG16, which exceeds 138~million trainable weights. In total, the proposed network contains only 341k parameters ($\sim$1.3~MB), making it over 70$\times$ smaller than ResNet-50 while achieving a 1.1\% higher accuracy on the 59-class SensiCut dataset. This balance of compactness and discriminative capability enables real-time inference on edge hardware, demonstrating that domain-specific lightweight design can outperform generic deep backbones when tailored to speckle-based classification tasks.

\section{Performance Evaluation}
\label{sec:evaluation}
\subsection{Experimental Setup}
Model training was carried out using the Adam optimizer with a learning rate schedule tuned for stable convergence. The loss function was categorical cross-entropy, appropriate for multi-class classification with mutually exclusive labels. A batch size of 64 was selected as a compromise between computational efficiency and gradient stability, and training was allowed to proceed for a maximum of 500 epochs. To prevent overfitting, an early stopping criterion was implemented with validation accuracy as the monitored metric, along with model checkpointing to retain the best-performing weights. The final model was selected at epoch 493, corresponding to a validation accuracy of 93.76\%.

Evaluation of the model was conducted using multiple performance metrics to provide a holistic view of classification ability. In addition to overall accuracy, Precision, recall, macro-averaged F1-score, and weighted F1-score are reported. Confusion matrices were generated for both fine-grained (59-class) predictions and grouped predictions at the nine-category and five-category family levels. These matrices allowed for detailed analysis of inter-class confusions and practical performance when families of materials are considered. Finally, inference speed was measured over the test set to assess real-time feasibility, recording both average per-sample latency and throughput.

\subsection{Performance on 59 Classes}
\begin{figure}[htbp]
    \centering
    \includegraphics[width=\columnwidth]{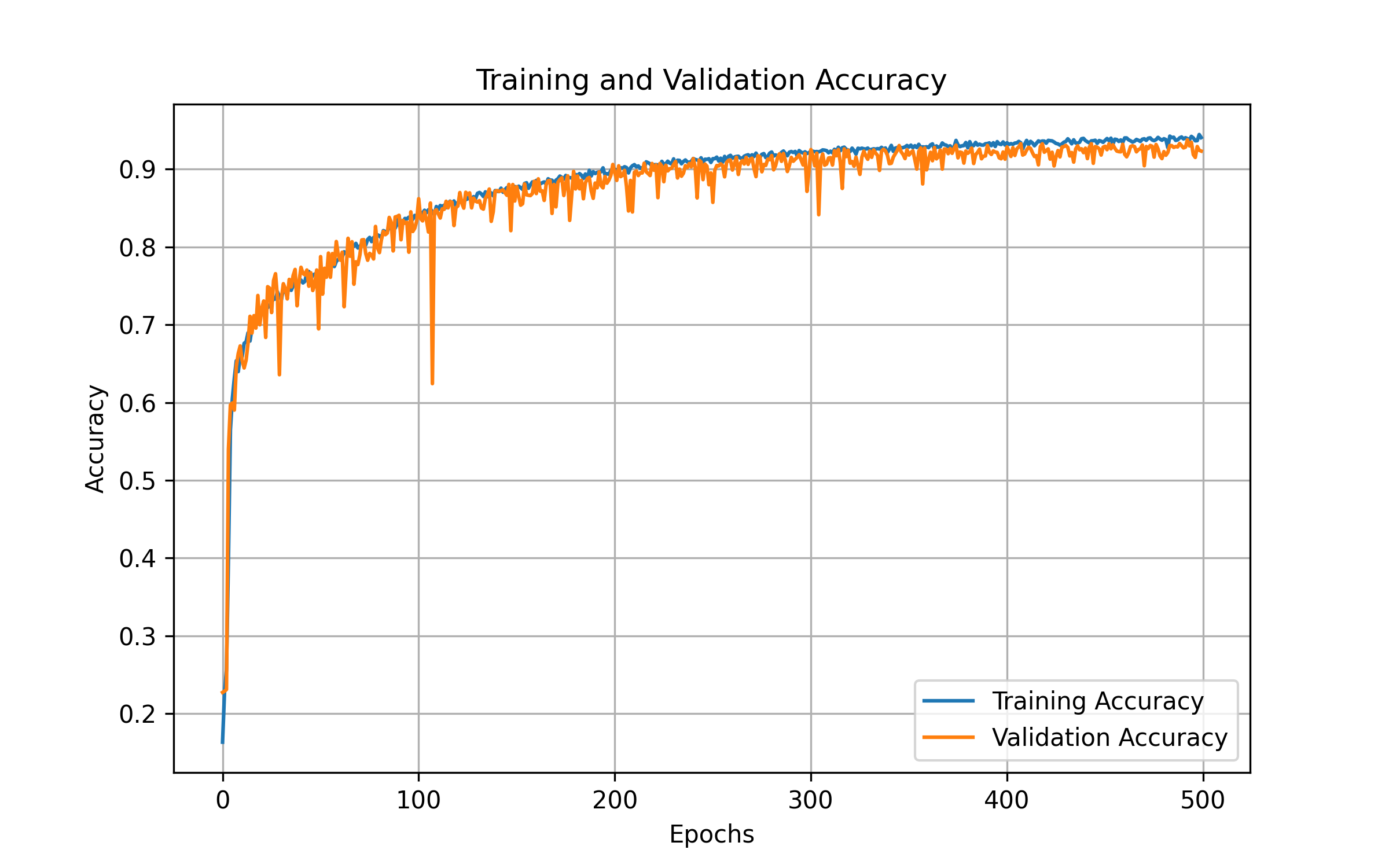}
\caption{Training and validation accuracy across epochs. The model converges stably, reaching a validation accuracy of 93.76\% at epoch 493.}
\label{fig:acccuracy}
\end{figure}

\begin{figure}[htbp]
    \centering
    \includegraphics[width=\columnwidth]{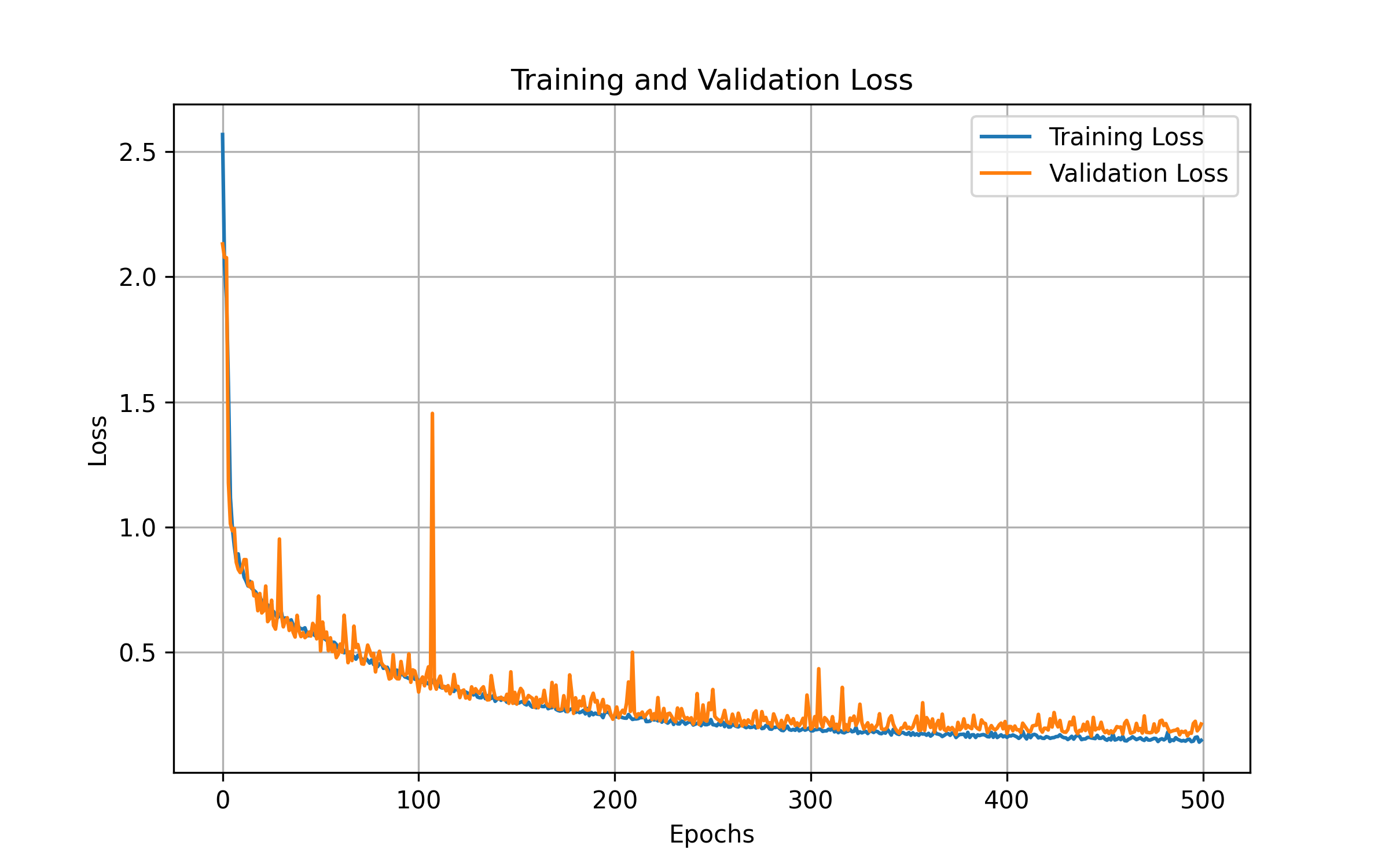}
\caption{Training and validation loss curves. The model shows no evidence of overfitting, confirming the effectiveness of augmentation and regularization.}
\label{fig:loss}
\end{figure}

\begin{figure}[htbp]
    \centering
    \includegraphics[width=\columnwidth]{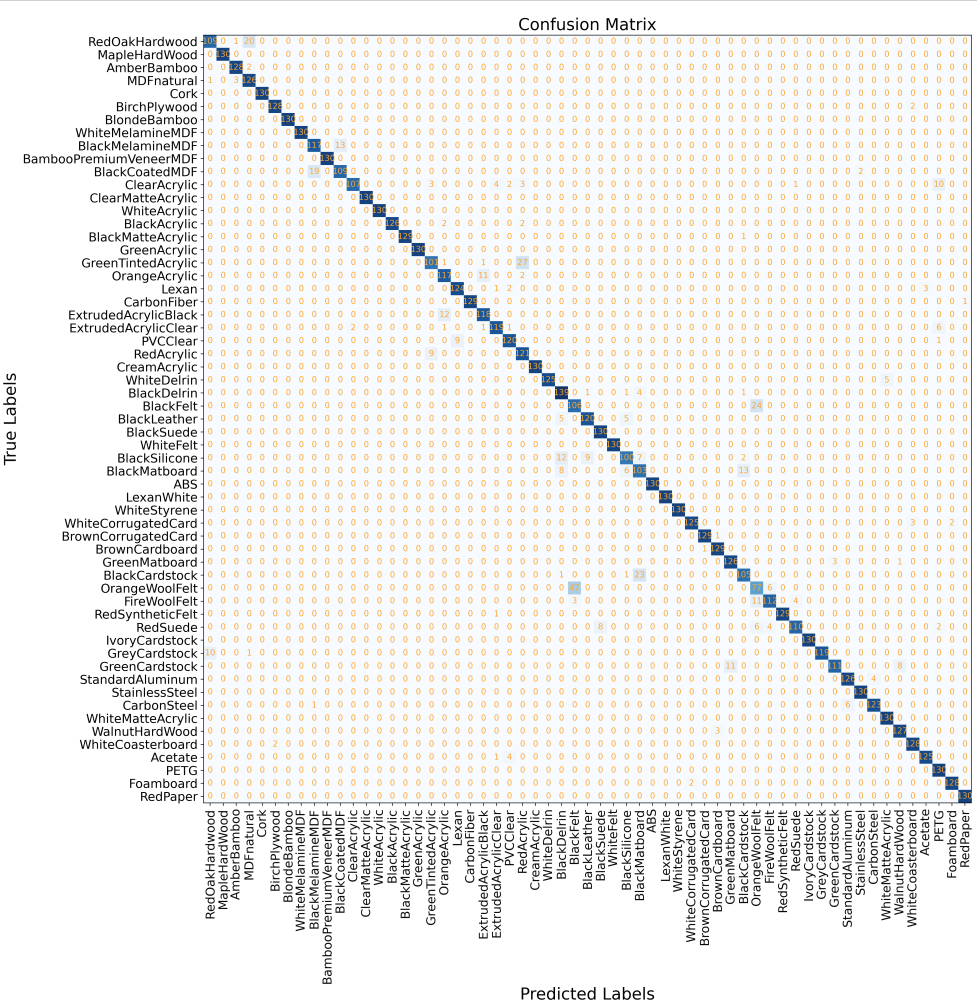}
\caption{Confusion matrix for the 59-class classification task. Strong diagonal dominance demonstrates high per-class accuracy, with misclassifications primarily occurring among similar materials such as hardwoods, tinted acrylics, and felts.}
\label{fig:cm59}
\end{figure}

On the complete 59-class SensiCut dataset, the proposed CNN achieved a test accuracy of 95.05\%. The macro F1-score was 0.951, and the weighted F1-score was also 0.951, confirming balanced performance across classes. Fig.~\ref{fig:acccuracy} and Fig.~\ref{fig:loss} show the training and validation accuracy and loss curves, demonstrating stable convergence without overfitting. The 59-class confusion matrix (Fig.~\ref{fig:cm59}) shows strong diagonal dominance, indicating that most classes were classified correctly. Misclassifications primarily occurred among visually similar categories, such as different hardwood variants, tinted acrylics, and textile felts, which is consistent with limitations reported in prior SensiCut studies.

\subsection{Grouped Categories}
\begin{figure}[htbp]
    \centering
    \includegraphics[width=\columnwidth]{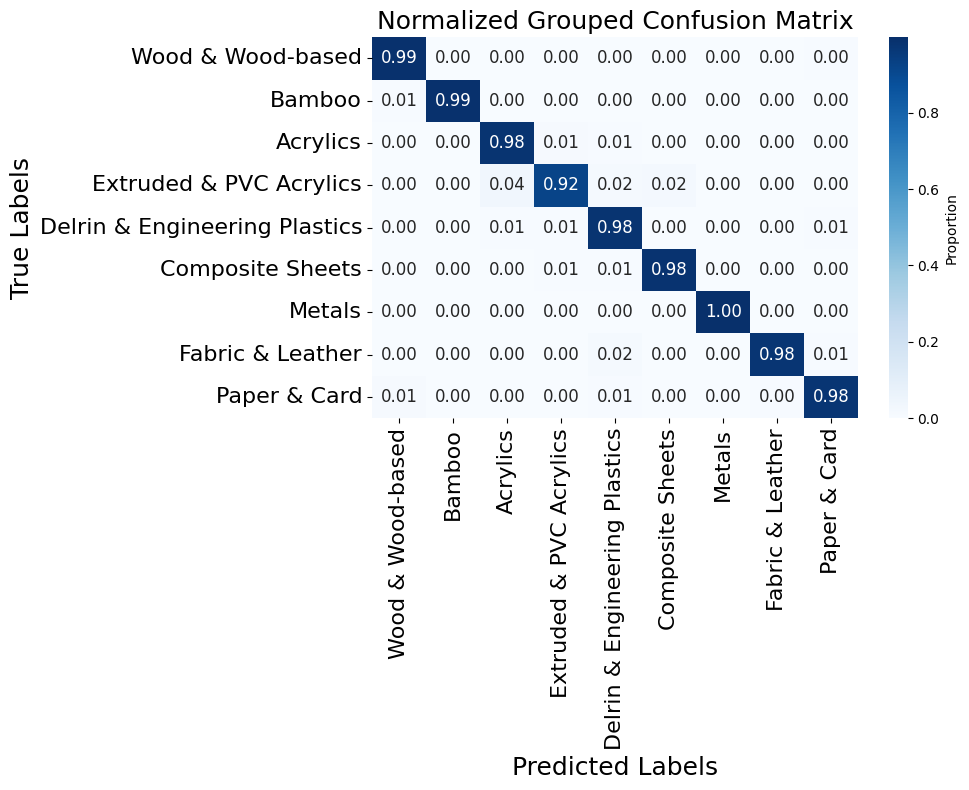}
\caption{Confusion matrix for the nine-family grouping.}
\label{fig:9GCM}
\end{figure}

\begin{figure}[htbp]
    \centering
    \includegraphics[width=\columnwidth]{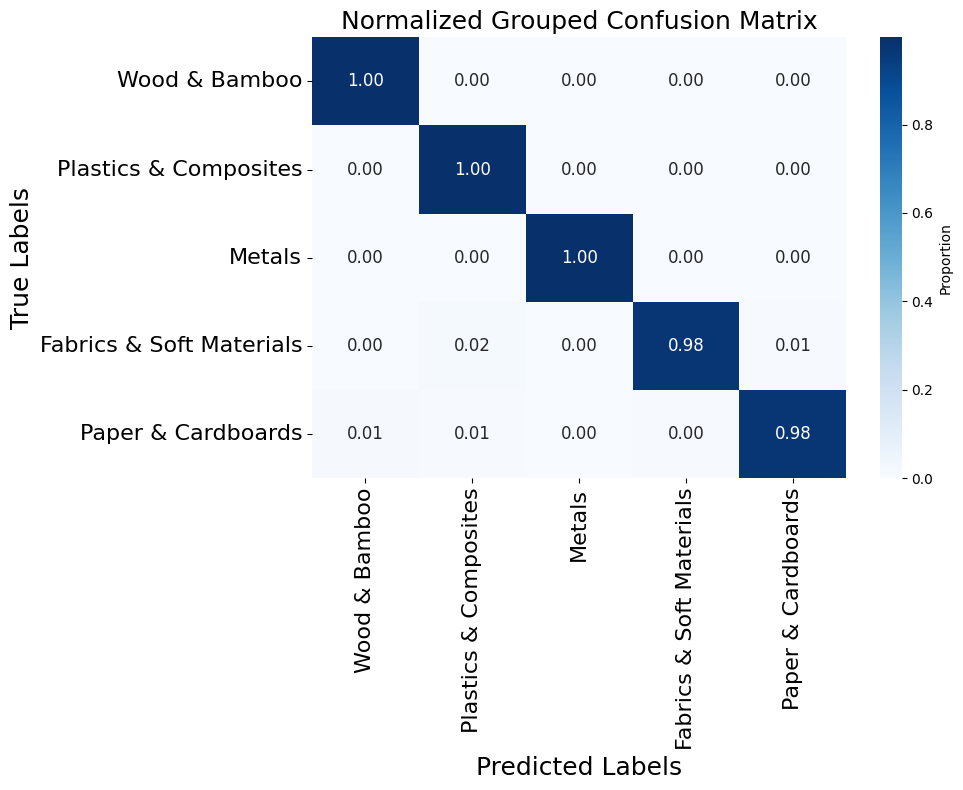}
\caption{Confusion matrix for the five-family grouping.}
\label{fig:5GCM}
\end{figure}

When the dataset was regrouped into nine high-level material families, the proposed model achieved recall above 0.92 for all categories, with most exceeding 0.98. This level of performance demonstrates the robustness of the classifier when materials are categorized into families with similar physical properties. At the coarser five-family grouping, recognition accuracy approached 99--100\%, underscoring the model's reliability for practical use. In real-world laser cutting, such family-level grouping is often sufficient, as materials within a family typically share similar cutting parameters such as power and speed. The confusion matrices for both the nine- and five-family scenarios are shown in Fig.~\ref{fig:9GCM} and Fig.~\ref{fig:5GCM}, respectively.

\subsection{Comparison with ResNet-50}
To contextualize the performance of the proposed CNN, a comparison was made with ResNet-50, which served as the backbone in the original SensiCut system. Table~\ref{tab:compare} summarizes the results. While ResNet-50 achieved 98.0\% accuracy on 30 classes, performance dropped to 93.96\% when extended to all 59 classes. In contrast, the porposed lightweight CNN achieved 95.05\% accuracy across the same 59 classes, representing a 1.1\% improvement while being over 70$\times$ smaller in parameter count. This highlights the advantage of a domain-tailored lightweight architecture in both scalability and efficiency.

\begin{table}[htbp]
\caption{Comparison of ResNet-50 and Proposed CNN on the SensiCut Dataset}
\begin{center}
\setlength{\tabcolsep}{3pt} % Default is usually 6pt. Reducing this tightens the table.
\begin{tabular}{|c|c|c|c|}
\hline
\textbf{Model} & \textbf{\textit{Parameters}} & \textbf{\textit{Classes}} & \textbf{\textit{Accuracy}} \\
\hline
ResNet-50 (30 classes) & 25M & 30 & 98.0\% \\
\hline
ResNet-50 (59 classes) & 25M & 59 & 93.96\% \\
\hline
Proposed CNN (59 classes) & 0.34M & 59 & \textbf{95.05\%} \\
\hline
\multicolumn{4}{l}{\textsuperscript{a} Proposed CNN achieves higher accuracy with $\sim$70$\times$ fewer parameters.}
\end{tabular}
\label{tab:compare}
\end{center}
\end{table}

\subsection{Efficiency}
The compact design of the proposed CNN also translates into practical efficiency gains. The model achieved an average inference time of 0.00339 seconds per sample, corresponding to approximately 295 images per second. With a memory footprint of only 1.3 MB, the model can be deployed on edge hardware such as Raspberry Pi or NVIDIA Jetson devices without modification. This level of efficiency ensures real-time classification capability, which is essential for integration into laser cutters where material recognition must occur seamlessly during operation.

\section{Discussion}
\label{sec:discussion}
The results demonstrate that a lightweight CNN can not only match but also surpass the performance of heavyweight backbones such as ResNet-50 for speckle-based material recognition. By relying solely on the green channel, preprocessing is simplified while preserving sufficient discriminative information, effectively reducing input redundancy and computational complexity. The strong family-level performance further indicates that the classifier is well suited for real-world deployment, where accurate identification of material families is sufficient to select appropriate laser power and speed presets.

The analysis also reveals that misclassifications are primarily concentrated among materials with highly similar surface scattering characteristics-most notably tinted acrylics and textile felts. These confusions are likely due to overlapping speckle statistics or limited training diversity for those specific subclasses. Despite these challenges, the model maintains high accuracy across diverse materials, confirming its generalization capability. Moreover, the inference speed of nearly 300 images per second demonstrates that speckle-based classification can feasibly be integrated into embedded laser cutter controllers without affecting operational throughput.

Overall, the discussion highlights that domain-specific network design and channel optimization can achieve state-of-the-art performance with minimal parameters. The combination of efficiency, interpretability, and scalability positions the proposed approach as a strong foundation for next-generation, material-aware laser cutting systems.

\section{Conclusion and Future Work}
\label{sec:conclusion}
This study presented a compact and efficient convolutional neural network (CNN) for speckle-based material classification in laser cutting systems. The proposed model achieves high accuracy (95.05\%) across all 59 material classes of the SensiCut dataset while maintaining a lightweight design of only 341k trainable parameters ($\sim$1.3~MB). Compared to the 25M-parameter ResNet-50 baseline, the proposed architecture delivers similar or superior performance while being approximately 70$\times$ smaller and capable of real-time inference at 295~images/s on edge devices such as the Raspberry~Pi and NVIDIA~Jetson platforms. These results demonstrate that domain-tailored, resource-efficient CNNs can enable practical, material-aware, and safety-oriented automation in laser cutting environments.

Beyond achieving fine-grained recognition, the model effectively generalizes when materials are grouped into nine and five high-level families, reaching near-perfect accuracy ($\geq$98\%). These family-level predictions directly correspond to laser cutter power and speed presets, bridging the gap between visual recognition and machine control. The proposed approach thus offers a scalable pathway toward intelligent fabrication systems that can autonomously adjust cutting parameters based on detected material type.

While the current results are promising, several directions can further enhance this research. First, extending the approach to multi-channel or multi-wavelength speckle inputs may improve discrimination among visually similar materials such as tinted acrylics and textile felts. Second, integration of lightweight attention mechanisms or hybrid CNN-Transformer architectures could enhance feature interpretability and robustness. Third, hardware-aware model optimization techniques such as pruning, quantization, and mixed-precision inference should be explored to achieve even lower latency and energy consumption on embedded laser cutter controllers. Finally, future work may expand beyond fabrication by adapting this framework to related domains such as agricultural sensing, fruit quality inspection, and robotic weed detection, where speckle-based surface analysis can offer fast, non-destructive assessment.

\bibliographystyle{IEEEtran}
\bibliography{references}

\vspace{12pt}

\end{document}